\documentclass[11pt]{article}

\usepackage{amssymb}
\usepackage{amsmath} 
\usepackage[final]{acl}

\usepackage{times}
\usepackage{latexsym}
\usepackage[T1]{fontenc}
\usepackage[utf8]{inputenc}
\usepackage{microtype}
\usepackage{inconsolata}
\usepackage{graphicx}
\usepackage{amsmath}
\usepackage{amssymb}
\usepackage{booktabs}
\usepackage{xcolor}
\usepackage{colortbl}
\PassOptionsToPackage{hyphens}{url}
\usepackage{hyperref}

\usepackage{multirow}
\usepackage{float}

\newcommand{\method}{\textsc{OISD}}
\newcommand{\sg}{\operatorname{sg}}

\newcommand{\maybeincludegraphics}[2][]{%
\IfFileExists{#2}{\includegraphics[#1]{#2}}{\fbox{\parbox[c][0.23\textheight][c]{0.92\linewidth}{\centering Missing figure file: \texttt{\detokenize{#2}}}}}%
}
\definecolor{ourscolor}{RGB}{235,232,255}
\usepackage[T1]{fontenc}

\usepackage[utf8]{inputenc}

\usepackage{microtype}

\usepackage{inconsolata}

\usepackage{graphicx}

\title{OISD: On-Policy Internal Self-Distillation of Language Models}

\author{
  \textbf{Xinyu Liu\textsuperscript{1}},
  \textbf{Darryl Cherian Jacob\textsuperscript{1}},
  \textbf{Yang Zhou\textsuperscript{1}},
  \textbf{Jindong Wang\textsuperscript{2}},
  \textbf{Pan He\textsuperscript{1}\textsuperscript{\textdagger}}
\\
  \textsuperscript{1}Auburn University \quad
  \textsuperscript{2}William \& Mary
\\
  \texttt{\{xil0004, dzj0055, yangzhou, pan.he\}@auburn.edu};
  \texttt{jdw@wm.edu}
\\\textsuperscript{\textdagger}Corresponding author
}

\begin{document}
\maketitle
\begin{abstract}

Recent reinforcement learning (RL) post-training approaches primarily optimize the final output policy using sparse outcome-level rewards, while largely overlooking predictive signals encoded in intermediate representations. In this paper, we  introduce a new paradigm called on-policy
internal self-distillation and propose the \textsc{OISD} framework, which improves reasoning by transferring on-policy predictive signals from the final layer to intermediate representations. During rollout and Group Relative Policy Optimization (GRPO) optimization, the final layer acts as both the policy and a detached internal teacher for selected intermediate layers, which are guided to align with it through two complementary mechanisms: logit alignment, which transfers high-level reasoning behaviors (\emph{how to think}), and attention alignment, which enforces consistent attention patterns (\emph{where to look}) from the final layer to the selected intermediate layer, both without requiring external privileged information. Our \textsc{OISD}, together with GRPO, employs signed advantage-weighted Jensen--Shannon alignment to distill informative intermediate representations while preserving policy consistency under a unified acting policy. Experimental results demonstrate the effectiveness of \textsc{OISD}, with substantial and consistent improvements over strong reasoning RL baselines across four mathematical reasoning tasks.

\end{abstract}

\section{Introduction}

\begin{figure}[t]
\centering
\includegraphics[width=\linewidth]{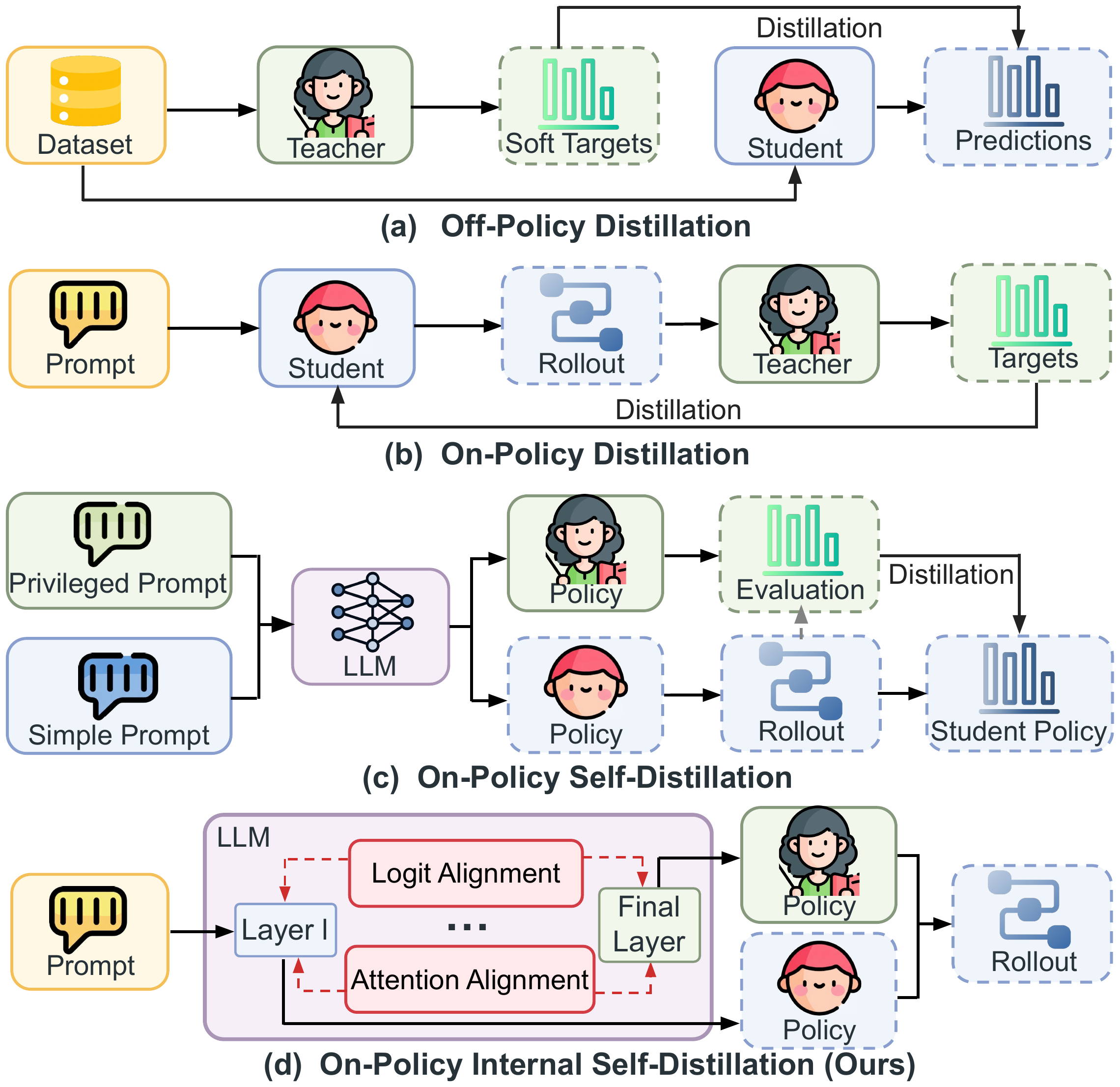}
\caption{Comparison of four distillation paradigms: (a) Off-Policy Distillation, (b) On-Policy Distillation, (c) On-Policy Self-Distillation, and (d) On-Policy Internal Self-Distillation (Ours). The comparison highlights differences in teacher–student roles, rollout generation, and the location where distillation is applied.}
\label{fig:methods-comparison}   
\end{figure}

Large language models (LLMs) have achieved remarkable advances in mathematical reasoning, coding, and complex instruction-following~\citep{yao2022react,wei2022chain}. A key driver of this progress is reasoning-oriented post-training, in which models are further optimized after pretraining and supervised fine-tuning to generate more accurate, coherent, and reliable reasoning trajectories~\citep{shao2024deepseekmath,guo2025deepseek}.

Recent reinforcement learning with verifiable rewards (RLVR), particularly Group Relative Policy Optimization (GRPO)-style training, has significantly improved reasoning-oriented post-training without requiring dense human supervision~\citep{shao2024deepseekmath,guo2025deepseek,zheng2025group,yu2026dapo}.
However, these approaches remain heavily final-output-centric, optimizing only the final acting policy with sparse outcome rewards while largely overlooking intermediate predictions, reasoning processes, and internal representations.

Knowledge distillation~\citep{hinton2015distilling,bengio2015scheduled} provides dense token-level supervision along the reasoning trajectory, complementing sparse outcome rewards. In classical off-policy distillation, the student learns from fixed teacher-generated traces, creating a distribution mismatch because the training trajectories come from the teacher rather than the student’s own generations~\citep{agarwal2024policy,xu2025speculative}. On-policy distillation reduces this gap by supervising student-generated rollouts directly~\citep{agarwal2024policy}, but most methods still rely on external teacher models. Recent on-policy self-distillation methods remove the external teacher by augmenting the same model with privileged traces, feedback, or auxiliary prompts~\citep{zhao2026self}. However, supervision still depends on externally injected privileged information rather than the model’s own internal computations.

Existing off-policy and on-policy distillation methods primarily focus on improving model accuracy and reasoning capability, with limited exploration of leveraging the model’s own internal mechanisms for distillation. However, prior studies~\citep{yang2025unveiling,zhu2025surveylatentreasoning} show that reasoning representations evolve systematically across layers: earlier layers preserve broader candidate predictions, intermediate layers integrate contextual and task-relevant information, and deeper layers progressively converge toward the final next-token decision. This hierarchical organization suggests that different layers serve distinct functional roles in the reasoning process. 

Meanwhile, the logit lens framework~\citep{nostalgebraist2020logitlens} provides initial insights by projecting intermediate representations into token space using the unembedding matrix. This reveals that the residual stream in Transformer architectures contains rich predictive information that evolves progressively across layers and modules~\citep{dai2022knowledge,gupta2025llms}. 
All these insights motivate leveraging the model’s own internal computations as an intrinsic source of supervision for distillation.

In this paper, we propose \textsc{OISD}, an on-policy internal self-distillation framework for LLM reasoning. During training, \textsc{OISD} keeps the final layer as the sole acting policy for rollout and GRPO optimization, while using it as a detached internal teacher. Following the logit lens framework~\citep{belrose2023eliciting}, a selected intermediate layer is projected into the token space to form an auxiliary internal student on the same on-policy rollout. The student learns from two complementary signals: logit alignment, which transfers high-level reasoning behaviors (\emph{how to think}), and attention alignment, which enforces consistent attention patterns (\emph{where to look}) from the final layer to the selected intermediate layer, both without requiring external privileged information. Such a formulation encourages the model to learn stronger intermediate representations. During inference, the teacher model is used as the final model. \textsc{OISD} introduces dense internal supervision without requiring an external teacher, privileged prompts, or multiple acting policies, motivating the term \emph{on-policy internal self-distillation}. A comparison between our OISD and other methods is shown in Figure~\ref{fig:methods-comparison}.

\subsection{Contributions}
Our contributions are summarized as follows:
\begin{itemize}
    \item We introduce a new paradigm called \emph{on-policy internal self-distillation} for reasoning RL that leverages the model’s own internal computations as supervision while preserving a single acting policy for rollout and optimization.
    
    \item We propose \textsc{OISD}, an on-policy internal self-distillation framework that uses the final layer as a detached internal teacher and an intermediate layer as an auxiliary internal student and transfers on-policy predictive signals from the final layer to intermediate representations.
    
    \item We develop two complementary internal alignment objectives: logit alignment for \emph{how to think} by transferring predictive beliefs, and attention alignment for \emph{where to look} by transferring evidence-routing behavior.
    
    \item We demonstrate substantial and consistent improvements over strong baselines across four mathematical reasoning benchmarks.
\end{itemize} We hope this research initiates the investigation of on-policy internal self-distillation and encourages future research in this direction.

\section{Related Work}
\noindent \textbf{Knowledge Distillation for Reasoning.}  Recent Reinforcement Learning from Human Feedback (RLHF) methods based on PPO provide a standard framework for preference optimization~\citep{schulman2017proximal,ouyang2022training}. Later, reasoning-oriented RL methods, such as GRPO, DeepSeek-R1, DAPO, GSPO, Qwen3, and RAIF, leverage verifiable rewards and group-based policy optimization to improve long context reasoning~\citep{shao2024deepseekmath,guo2025deepseek,zheng2025group,yang2025qwen3,qin2026incentivizing,yu2026dapo}. However, supervision remains largely sparse, outcome-level, and focused on the final acting policy.

To provide denser reasoning supervision, prior work~\citep{hinton2015distilling,bengio2015scheduled} often distills teacher distributions or reasoning traces into the student. Yet off-policy distillation introduces a training--inference mismatch, since the student is trained on teacher-generated trajectories but must reason from its own internal states during inference.

On-policy distillation reduces this mismatch by generating rollouts from the student and using the teacher’s per-token log-probabilities as dense supervision on states actually visited by the student~\citep{agarwal2024policy,xu2025speculative}. 
More recent self-distillation methods reduce reliance on external teachers by enabling the model to supervise itself using privileged information or richer feedback~\citep{agarwal2024policy,hubotter2026reinforcement,shenfeld2026self}. For example, OPSD~\cite{zhao2026self} leverages privileged solutions, SDPO~\cite{hubotter2026reinforcement} exploits environment feedback, OPCD~\citep{ye2026policy} distills context-conditioned behavior into model weights, and CODI~\cite{shen2025codi} compresses explicit reasoning into continuous representations. 

Our method places the teacher entirely within the model itself: the detached final layer supervises an intermediate layer on the same on-policy rollout while remaining the sole acting policy.

\noindent \textbf{Internal Signals for Reasoning.} Early research efforts, such as deeply-supervised nets (DSN)~\citep{lee2015deeply}, set a broader optimization precedent for attaching auxiliary losses to intermediate representations. More recently, Layerwise readouts show that intermediate Transformer representations already contain meaningful predictive structure before the final layer is reached. Logit-lens and tuned-lens analyses decode hidden states into vocabulary distributions, making prediction trajectories across layers directly observable~\citep{nostalgebraist2020logitlens,belrose2023eliciting}. 

Later, DoLa uses contrasts between layers during decoding, while LayerSkip trains intermediate exits for efficient inference and self-speculative decoding~\citep{chuang2024dola,elhoushi2024layerskip}. In particular, BuPO pioneered the investigation in optimizing fine-grained internal layer policies to guide the overall policy of the language model during the initial training stages~\cite{tan2025bottom}. However, the authors observed that excessive alignment can lead to performance collapse, suggesting that intermediate policy optimization should be applied only for limited training iterations. Other approaches use layerwise signals for analysis, decoding, early exit, or inference efficiency rather than as auxiliary supervision~\cite{kapadia2026leap}.

Reasoning supervision can also arise from the reasoning process itself, not only from final answers. CODI compresses explicit reasoning into continuous internal representations through self-distillation~\citep{shen2025codi}, while step-level verification methods study supervision over intermediate reasoning steps~\citep{lightman2024let,zheng2025processbench,song2025prmbench}. Attention provides another process-level signal by revealing how information is routed during token prediction. For example, MoLSAKI transfers stepwise attention from a teacher to a student during reasoning distillation~\citep{chen2025improving}, and RAL directly optimizes internal attention distributions through policy-gradient training~\citep{li2026reinforced}.

Our \textsc{OISD} develops two complementary internal alignment objectives in an on-policy setting: logit alignment supervises \emph{how to think}, attention alignment supervises \emph{where to look}, and both are distilled from the final layer to an intermediate layer on the same rollout.

\section{On-Policy Internal Self-Distillation }
\label{sec:method}

In this section, we present \textsc{OISD}, an on-policy internal self-distillation framework that uses the final layer as a detached internal teacher and an intermediate layer as an auxiliary internal student during reasoning RL. We further introduce two complementary internal alignment objectives for supervising intermediate reasoning representations.

\subsection{Transformer Depth as a Vertical Reasoning Axis}

CoT reasoning exposes intermediate computation through additional reasoning tokens, expanding reasoning \emph{horizontally} along the temporal dimension~\citep{wei2022chain}. More recent latent reasoning methods further extend this paradigm by evolving hidden states or recurrent activations across reasoning steps rather than fully expressing reasoning in natural language~\citep{hao2024training,shen2025codi}. Following~\citep{zhu2025surveylatentreasoning}, these approaches primarily expand reasoning capacity through hidden-state evolution over time.

\textsc{OISD} instead explores a complementary \emph{vertical} reasoning paradigm within a standard decoder-only Transformer consisting of $L$ stacked layers. Rather than extending reasoning through additional tokens, recurrent activations, or latent rollout steps, \textsc{OISD} leverages the intermediate computations already formed across model depth. The final layer supervises intermediate layers during the same forward pass, allowing reasoning signals to propagate \textit{vertically} through the network without introducing latent tokens or modifying the decoding procedure.

To illustrate the vertical computation within autoregressive generation,
following~\citep{zhang2025reinforcement,tan2025bottom}, we formalize one decoding step. Let
$\mathbf{x}=[x_1,\ldots,x_M]$ denote the original input prompt, and let
$\mathbf{y}_{1:t-1}=[y_1,\ldots,y_{t-1}]$ denote the response prefix generated before step $t$, with
$\mathbf{y}_{1:0}=\emptyset$ for the first decoding step. At generation step $t$, the model receives the full decoding context
\begin{equation}
  \mathbf{c}_t = [\mathbf{x};\mathbf{y}_{1:t-1}], \qquad t=1,\ldots,T .
\end{equation}
The model maps this context into initial hidden states
$\mathbf{H}^{0}_{t}\in\mathbb{R}^{|\mathbf{c}_t|\times d_{\text{model}}}$ using the embedding matrix
$\mathbf{E}\in\mathbb{R}^{N\times d_{\text{model}}}$. Each Transformer layer then updates the residual stream over all context positions through multi-head self-attention
(MHSA) and feed-forward network (FFN) transformations:
\begin{equation}
\begin{aligned}
\mathbf{A}^l_{t} &= \mathrm{MHSA}\!\left(\mathrm{LN}(\mathbf{H}^{l-1}_{t})\right), \\
\bar{\mathbf{H}}^l_{t} &= \mathbf{H}^{l-1}_{t} + \mathbf{A}^l_{t}, \\
\mathbf{F}^l_{t} &= \mathrm{FFN}\!\left(\mathrm{LN}(\bar{\mathbf{H}}^l_{t})\right), \\
\mathbf{H}^l_{t} &= \bar{\mathbf{H}}^l_{t} + \mathbf{F}^l_{t},
\end{aligned}
\label{eq:transformer-layer}
\end{equation}
for $l=1,\ldots,L$, where $\mathrm{LN}(\cdot)$ denotes layer normalization~\citep{ba2016layer}. 

This formulation separates two axes of computation. Along the autoregressive sequence axis, the context grows from
$\mathbf{c}_t=[\mathbf{x};\mathbf{y}_{1:t-1}]$ to
$\mathbf{c}_{t+1}=[\mathbf{x};\mathbf{y}_{1:t}]$ after sampling $y_t$. Along the layer axis, the model performs vertical refinement, progressively transforming
representations from shallow to deep layers. After $L$ layers, the next-token policy is obtained from the final hidden state:
\begin{equation}
\mathbf{p}_{t}
=
\mathrm{Softmax}
\left(
\mathrm{LN}(\mathbf{h}^{L}_{t})
\mathbf{E}_{\mathrm{u}}^\top
\right),
\quad
y_t \sim \mathbf{p}_{t},
\end{equation}
where $\mathbf{h}^{L}_{t}\in\mathbb{R}^{d_{\text{model}}}$ is the final-layer hidden state at the last position of context $\mathbf{c}_t$, $\mathbf{E}
_{\mathrm{u}}\in\mathbb{R}^{N\times d_{\text{model}}}$ is the unembedding matrix, and $\mathbf{p}_{t}\in\mathbb{R}^{N}$ denotes the distribution over the next generated
token $y_t$. The hidden states at layer $l$ can be recursively expanded as
\begin{equation}
\mathbf{H}^{l}_{t}
=
\mathbf{H}^{0}_{t}
+
\sum_{i=1}^{l}\mathbf{A}^{i}_{t}
+
\sum_{i=1}^{l}\mathbf{F}^{i}_{t},
\end{equation}
showing that each layer accumulates transformations from all preceding layers. 

\begin{figure}[t]
\centering
\maybeincludegraphics[width=\linewidth]{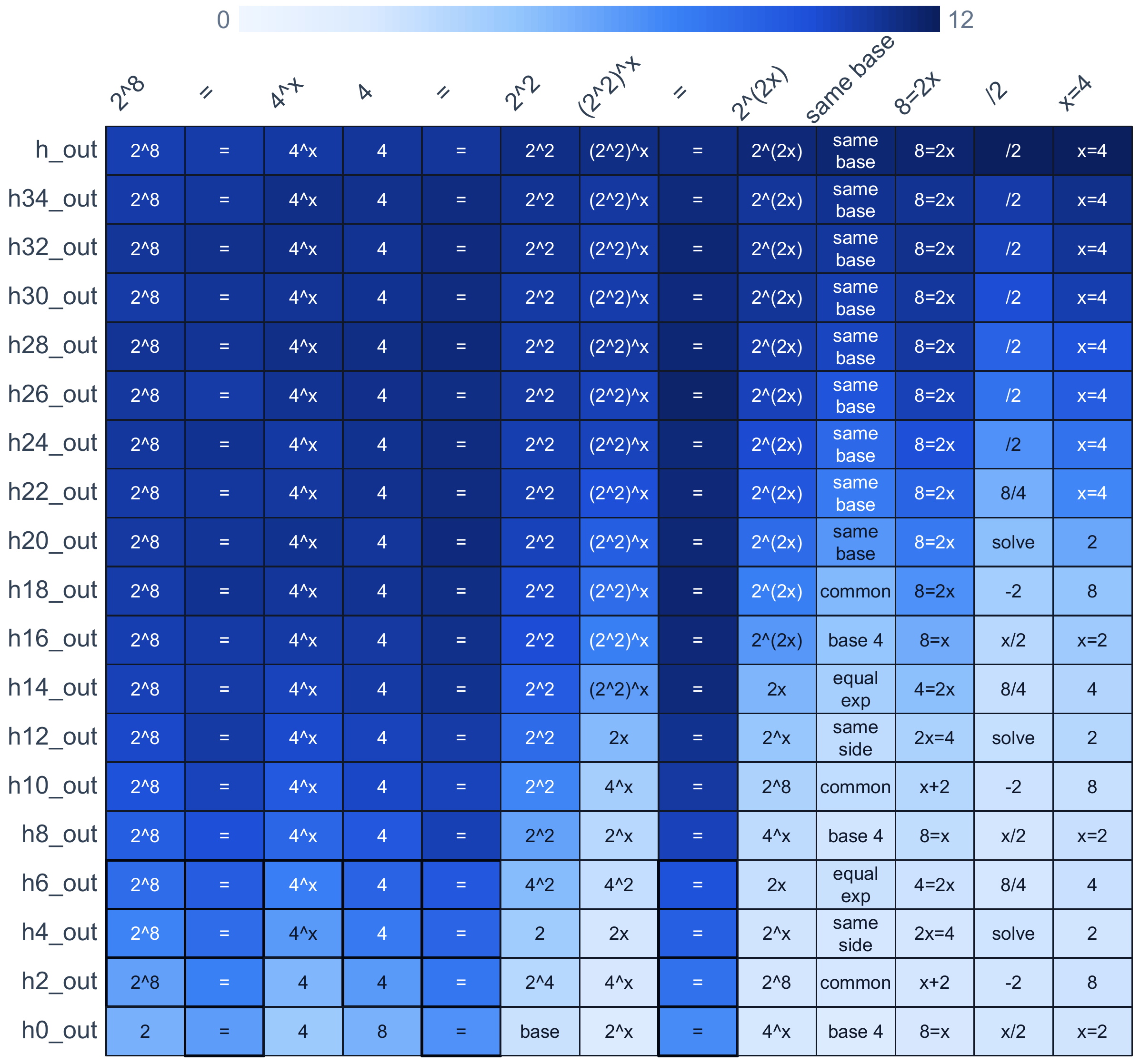}
\caption{Layerwise logit-lens visualization of a reasoning trajectory of Qwen3-4B. Columns denote generated reasoning tokens and rows denote Transformer layers. Each cell shows the top token from the projected hidden state $\mathbf{p}^{l}_{t}$, with darker color indicating higher logits. The visualization reveals progressive refinement, where local symbols emerge in early layers and multi-step algebraic reasoning stabilizes in deeper layers.}
\label{fig:layerwise-logit-lens}
\end{figure}

For each intermediate hidden state at layer i,  we can then project it back into token space using the same
unembedding matrix:
\begin{equation}
\mathbf{p}^{l}_{t}
=
\mathrm{Softmax}
\left(
\mathrm{LN}(\mathbf{h}^{l}_{t})
\mathbf{E}_{\mathrm{u}}^\top
\right),
\end{equation}
where $\mathbf{p}^{l}_{t}$ is the layer-$l$ next-token distribution induced by the hidden state $\mathbf{h}^{l}_{t}$. This projection leads to a provisional policy $y_t^l \sim \mathbf{p}_{t}^l$. Figure~\ref{fig:layerwise-logit-lens} visualizes the logit lens across layers.

\subsection{Guided Shallow Students, Stronger Deep Teachers}

Rather than treating intermediate states as transient features, \textsc{OISD} attaches auxiliary losses to a fixed layer $l<L$, treating $\mathbf{h}^{l}_{t}$ as a shallow
student and $\mathbf{h}^{L}_{t}$ as the deep teacher within the same forward pass. The teacher supervises the student through its token distribution and attention pattern, capturing both the belief reached by the model and the context used to form it. Because the student and teacher are induced from the same model at different depths, \textsc{OISD} uses the deep layer to guide better shallow representations; improving these early representations then strengthens the later computation and final-layer teacher.

\subsubsection{How to Think: Reward-Conditioned Logit Alignment} 

The readout $y_t^{l}\sim\mathbf{p}_{t}^{l}$ reveals the provisional next-token belief encoded at layer $l$, which we interpret as the shallow student's intermediate thinking process before deeper vertical refinement. \textsc{OISD} transfers
token-level beliefs from the deep teacher to the shadow student by comparing their logit-lens distributions. With temperature $\tau$, the student and teacher distributions
are
\begin{equation}
\begin{aligned}
\mathbf{p}^{l,\tau}_{t}
  & =
  \mathrm{Softmax}
  \left(
  \mathrm{LN}(\mathbf{h}^{l}_{t})\mathbf{E}_{\mathrm{u}}^\top/\tau
  \right), \\
  \mathbf{q}^{L,\tau}_{t}
  & = 
  \mathrm{Softmax}
  \left(
  \mathrm{LN}(\mathbf{h}^{L}_{t})\mathbf{E}_{\mathrm{u}}^\top/\tau
  \right).
\end{aligned}
\end{equation}
Their discrepancy can be measured by a bounded divergence measure. Formally, our logit-alignment loss at layer $l$ is defined as
\begin{equation}
  \mathcal{L}_{\mathrm{think}}^{(l)}
  =
  \mathbb{E}_{t}
  \left[
  A_t \operatorname{JS}
  \left(
  \mathbf{p}^{l,\tau}_{t},
\operatorname{sg}(\mathbf{q}^{L,\tau}_{t})
  \right) 
  \right],
  \label{eq:logit-loss}
\end{equation}
where $A_t$ is the assigned sequence-level GRPO advantage, $\operatorname{JS}$ denotes the Jensen--Shannon divergence~\cite{menendez1997jensen}, and $\operatorname{sg}(\cdot)$ denotes stop-gradient, which treats the teacher distribution as a fixed target. We weight the divergence by $A_t$: high-advantage tokens pull the shallow student toward the deep teacher, while low-advantage tokens push it away from the teacher to avoid reinforcing poor continuations. We use the clipped advantage $A_t=\operatorname{clip}(A_t,-c,c)$, with a constant $c$.


\subsubsection{Where to Look: Reward-Conditioned Attention Alignment}

At each generation step, an LLM attends to selected parts of the prompt and previously generated tokens, such as the problem statement, earlier intermediate results, or nearby reasoning steps. \textsc{OISD} aligns attention patterns, using final-layer attention to guide the shallow student toward the evidence-routing behavior before producing the next token.

For layer $l$ and head $h$, let
$\mathbf{Q}^{l}_{h},\mathbf{K}^{l}_{h}\in\mathbb{R}^{|\mathbf{c}_t|\times d_h}$
denote the query and key matrices computed from the current context
$\mathbf{c}_t=[\mathbf{x};\mathbf{y}_{1:t-1}]$. Let
$\mathbf{q}^{l}_{t,h}$ be the query vector at the last position of $\mathbf{c}_t$, used to predict $y_t$, and let
$\mathbf{k}^{l}_{j,h}$ be the key vector at context position $j$. The causal key set
$\mathcal{C}_t=\{1,\ldots,|\mathbf{c}_t|\}$ indexes all prompt tokens and previously generated tokens available before $y_t$ is produced. The attention weight between
position $t$ and any preceding position $j$ ($j<t$) at layer $l$ is computed as
\begin{equation}
a^{l}_{t,h}(j)
=
\frac{
\exp\left((\mathbf{q}^{l}_{t,h})^\top\mathbf{k}^{l}_{j,h}/\sqrt{d_h}\right)
}{
\sum_{j'\in\mathcal{C}_t}
\exp\left((\mathbf{q}^{l}_{t,h})^\top\mathbf{k}^{l}_{j',h}/\sqrt{d_h}\right)
}.
\end{equation}

To reduce memory cost on long responses, we compute the loss only for sampled decoding steps $t$ and sampled causal keys
$\tilde{\mathcal{C}}_t\subseteq\mathcal{C}_t$. The sampled key set combines a recent local window with strided global positions, so it covers both nearby reasoning tokens
and distant prompt evidence. We renormalize attention over this sampled set:
\begin{equation}
\tilde{a}^{l}_{t,h}(j)
=
\frac{a^{l}_{t,h}(j)}
{\sum_{k\in\tilde{\mathcal{C}}_t}a^{l}_{t,h}(k)},
\qquad j\in\tilde{\mathcal{C}}_t.
\end{equation} Our attention-alignment loss at layer $l$ is defined as 
\begin{equation}
  \mathcal{L}_{\mathrm{attn}}^{(l)}
  =
  \mathbb{E}_{t}
  \left[
  A_t  \frac{1}{H}\sum_{h=1}^{H}
\operatorname{JS} \left(
\tilde{a}_{t,h}^{l},
\sg\left(\tilde{a}_{t,h}^{L}\right)
\right) \right]
  \label{eq:logit-loss}
\end{equation} where $H$ denotes the number of heads in transformers. Positive advantages encourage the shadow layer to follow the final layer's evidence use, while negative advantages reduce agreement with attention patterns with low-reward continuations.
  


\subsection{The OISD Training Objective}



We adopt GRPO~\citep{shao2024deepseekmath} as the base training signal. For each GRPO rollout batch
$\mathcal{B}=\{(\mathbf{x}_i,\mathbf{y}_i,A^i_{1:T_i})\}_{i=1}^{G}$, \textsc{OISD} computes both the final-layer RL loss and the internal student-teacher losses on
the same sampled trajectories:
\begin{equation}
 \begin{aligned}
\mathcal{L}_{\mathrm{OISD}}(\mathcal{B})
& =
\mathcal{L}_{\mathrm{GRPO}}(\mathcal{B})
+
\lambda_{\mathrm{think}}
\frac{1}{G}\sum_{g=1}^{G}
\mathcal{L}_{\mathrm{think}}^{(l,g)}
\\
&+  \lambda_{\mathrm{attn}}
\frac{1}{G}\sum_{g=1}^{G}
\mathcal{L}_{\mathrm{attn}}^{(l,g)}.
\label{eq:overall}
\end{aligned}   
\end{equation} Here, $\mathcal{L}_{\mathrm{GRPO}}(\mathcal{B})$ is the final-layer loss of GRPO. The auxiliary losses
$\mathcal{L}_{\mathrm{think}}^{(l,g)}$ and $\mathcal{L}_{\mathrm{attn}}^{(l,g)}$ are computed on the same $g$-th GRPO rollout. The coefficients $\lambda_{\mathrm{think}}$ and $\lambda_{\mathrm{attn}}$ control the strengths of belief
and attention transfer. We set $\lambda_{\mathrm{think}}=1$ and $\lambda_{\mathrm{attn}}=0.1$. For layer selection $l$, we follow~\cite{tan2025bottom} and choose the last FFN layer with a positive exploration signal. Details are provided in the experiment section and the appendix.

\section{Experiment}

\subsection{Experimental Setup}

\begin{table*}[t]
\centering
\caption{Avg@K results on AMC23, MATH500, AIME24 and AIME25.}
\label{tab:main_math_results}
\footnotesize
\renewcommand{\arraystretch}{1.15}
\setlength{\tabcolsep}{3.5pt}
\begin{tabular}{lccccc}
\toprule\toprule
\textbf{Methods} &
\begin{tabular}[c]{@{}c@{}}\textbf{AMC23}\textbf{ (Avg@16)}\end{tabular} &
\begin{tabular}[c]{@{}c@{}}\textbf{MATH500}\textbf{ (Avg@16)}\end{tabular} &
\begin{tabular}[c]{@{}c@{}}\textbf{AIME24}\textbf{ (Avg@32)}\end{tabular} &
\begin{tabular}[c]{@{}c@{}}\textbf{AIME25}\textbf{ (Avg@32)}\end{tabular} &
\textbf{Avg.} \\
\midrule
\textit{Qwen3-4B} & & & & & \\
Vanilla & 67.66 & 80.29 & 23.20 & 18.60 & 47.44 \\
PPO & 77.03 & 83.64 & 32.60 & 27.60 & 55.22 \\
Reinforce++ & 63.44 & 80.63 & 17.40 & 18.65 & 45.03 \\
RLOO & 77.66 & 82.73 & 30.83 & 24.79 & 54.00 \\
GRPO & 76.88 & 82.41 & 32.19 & 28.85 & 55.08 \\
BuPO & 81.09 & 84.90 & 36.88 & 31.15 & 58.51 \\
\rowcolor{ourscolor}
\method\ (ours) & \textbf{84.38} & \textbf{90.99} & \textbf{47.29} & \textbf{36.35} & \textbf{64.75} \\
\midrule
\textit{Qwen3-8B} & & & & & \\
Vanilla & 67.34 & 80.46 & 26.98 & 19.17 & 48.49 \\
GRPO & 85.94 & 88.05 & 49.48 & 33.54 & 64.23 \\
BuPO & 89.22 & 87.76 & \textbf{54.06} & 34.38 & 66.36 \\
\rowcolor{ourscolor}
\method\ (ours) & \textbf{90.62} & \textbf{89.38} & 52.40 & \textbf{34.48} & \textbf{66.72} \\
\midrule
\textit{OctoThinker-3B-Long-Base} & & & & & \\
Vanilla & 1.24 & 5.26 & 0.21 & 0.00 & 1.68 \\
GRPO & 27.50 & 46.07 & 0.63 & 0.10 & 18.58 \\
BuPO & 27.50 & \textbf{49.79} & 0.63 & 0.42 & 19.59 \\
\rowcolor{ourscolor}
\method\ (ours) & \textbf{27.81} & 49.41 & \textbf{1.98} & 0.42 & \textbf{19.91} \\
\midrule
\textit{OctoThinker-8B-Long-Base} & & & & & \\
Vanilla & 4.53 & 9.84 & 0.52 & 0.10 & 3.75 \\
GRPO & 34.84 & 56.89 & 2.50 & 2.19 & 24.11 \\
BuPO & 37.66 & 62.05 & 4.69 & 6.77 & 27.79 \\
\rowcolor{ourscolor}
\method\ (ours) & \textbf{39.38} & \textbf{66.19} & \textbf{7.29} & \textbf{7.50} & \textbf{30.09} \\
\bottomrule\bottomrule
\end{tabular}
\end{table*}

\begin{figure*}[h]
\centering
\maybeincludegraphics[width=\textwidth]{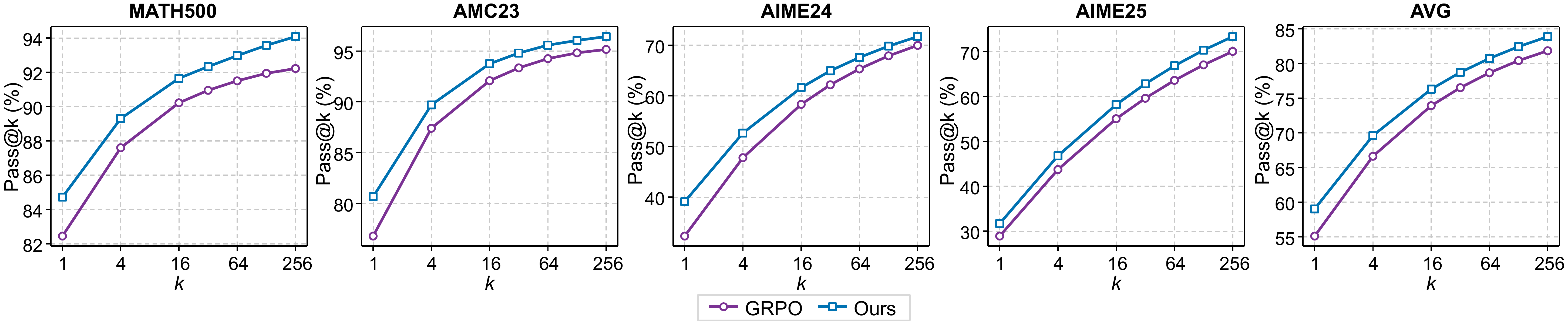}
\caption{Pass@$k$ comparison between GRPO and \method\ using Qwen3-4B.}
\label{fig:passk-performance}
\end{figure*}

\paragraph{Models and Baselines.}
We evaluate \textsc{OISD} on Qwen3-4B, Qwen3-8B, OctoThinker-3B-Long-Base, and OctoThinker-8B-Long-Base. Qwen3-4B~\citep{yang2025qwen3} is trained on \texttt{deepmath-5k} under the same setting as others. Qwen3-8B tests scaling within the
Qwen3 family, while the OctoThinker models evaluate transfer to a different model family. We compare \textsc{OISD} against the corresponding
Vanilla, PPO, Reinforce++, RLOO, GRPO, and BuPO~\citep{tan2025bottom}. GRPO serves as the final-layer-only RL baseline, while BuPO represents direct internal-policy optimization.  For layer selection $l$ for both $\mathcal{L}_{\mathrm{think}}^l$ and $\mathcal{L}_{\mathrm{attn}}^l$, we follow~\citep{tan2025bottom}, setting $l=6$ for Qwen3-4B and Qwen3-8B, $l=27$ for Llama-OctoThinker-3B-Base, and $l=31$ for Llama-OctoThinker-8B-Base.

\noindent \textbf{Evaluation benchmarks and metrics.}
The evaluation benchmarks include AMC23~\citep{AMC23}, MATH500~\citep{lightman2024let}, AIME24, and AIME25~\citep{AIME24,AIME25}. We
report Avg@16 for AMC23 and MATH500, Avg@32 for AIME24 and AIME25, and the mean across all four benchmarks. Avg@K denotes average accuracy over $K$ sampled responses. We also report Pass@K, which measures whether at least one of the $K$ sampled responses is correct under a fixed-sample evaluation protocol: $\mathrm{Pass@K} := \mathbb{E}_{x \sim D}\left[
1 - \frac{\binom{n-c}{K}}{\binom{n}{K}}
\right]$,
where $c$ is the number of correct completions among $n$ generated responses. To reduce evaluation variance, we use $n=300$ following~\cite{tan2025bottom}.






\begin{figure*}[t]
\centering
\includegraphics[width=\linewidth]{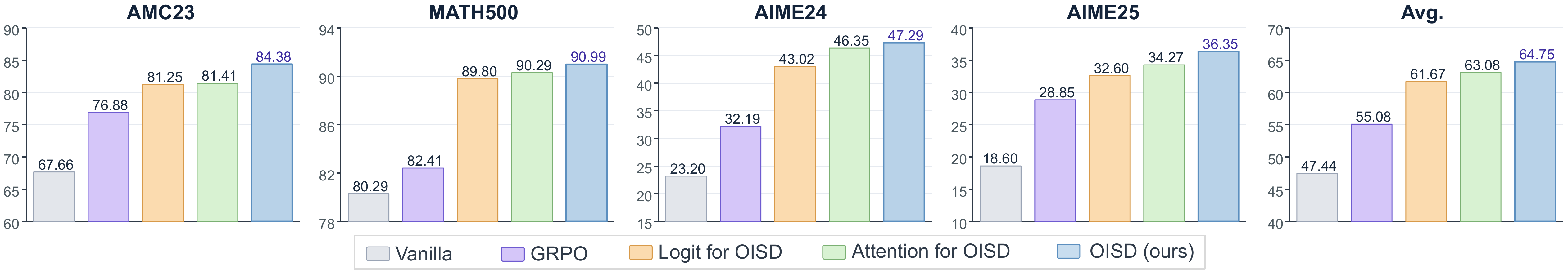}
\caption{
Effect of logit and attention alignment on Qwen3-4B. We compare Vanilla, GRPO, \method\ with logit alignment only, \method\ with attention alignment only, and the full \method\ on AMC23, MATH500, AIME24, AIME25 with the overall average.
}
\label{fig:component-ablation-heatmap}
\end{figure*}

\subsection{Main Results}

\noindent \textbf{Avg@K Performance.} Table~\ref{tab:main_math_results} reports the main Avg@$k$ results on AMC23, MATH500, AIME24, and AIME25. Using the Qwen3-4B backbone, \textsc{OISD} achieves the best average score of 64.75, outperforming GRPO by 9.67 points and BuPO by 6.24 points. The gains are especially strong on AIME24 and AIME25, with improvements of 14.31 and 9.21 points over GRPO, respectively. AMC23 and MATH500 also improve from 76.88 and 82.41 to 84.38 and 90.99. At the larger Qwen3-8B scale, \method\ attains the best overall average of 66.72 and remains competitive on the more challenging AIME benchmarks.

\noindent \textbf{Pass@K Performance.}
Figure~\ref{fig:passk-performance} demonstrates that 
increasing $K$ improves the pass rate for both GRPO and \method.
\method\ consistently achieves higher average Pass@$K$ across all evaluated settings, suggesting that OISD improves not only average accuracy but also the overall quality of the sampled solution set.

\begin{table}[t]
\centering
\caption{\footnotesize Robustness of \method\ across three independent Qwen3-4B training runs with seeds $1$, $42$, and $123$. AMC23 and MATH500 report Avg@16, while AIME24 and AIME25 report Avg@32.}
\label{tab:seed-robustness}
\footnotesize
\setlength{\tabcolsep}{3.2pt}
\renewcommand{\arraystretch}{1.1}
\resizebox{\linewidth}{!}{%
\begin{tabular}{lccccc}
\toprule
\textbf{Seed / Statistic} & \textbf{AMC23} & \textbf{MATH500} & \textbf{AIME24} & \textbf{AIME25} & \textbf{Avg.} \\
\midrule
Seed $1$   & 84.38 & 90.99 & 47.29 & 36.35 & 64.75 \\
Seed $42$  & 84.06 & 91.14 & 46.04 & 34.79 & 64.01 \\
Seed $123$ & 85.16 & 90.65 & 46.56 & 35.42 & 64.45 \\
\midrule
\rowcolor{ourscolor}
Mean & 84.53 & 90.93 & 46.63 & 35.52 & 64.40 \\
Std & 0.57 & 0.25 & 0.63 & 0.78 & 0.37 \\
$95\%$ CI & $\pm$1.41 & $\pm$0.62 & $\pm$1.56 & $\pm$1.95 & $\pm$0.93 \\
\bottomrule
\end{tabular}
}
\end{table}

\noindent \textbf{Robustness across Random Seeds.}
Since RLVR post-training is known to be noisy and the AIME benchmarks are small, we assess robustness across random seeds by conducting two additional Qwen3-4B runs with seeds $42$ and $123$, complementing the original run with seed $1$. All three runs use identical training data, hyperparameters, compute budgets, and experimental configurations. Table~\ref{tab:seed-robustness} reports the per-seed scores together with the mean, standard deviation, and confidence interval. \method\ achieves an average score of $64.40\pm0.37$ (mean $\pm$ standard deviation) across the three seeds, with a $95\%$ confidence interval of $[63.47, 65.33]$. The variation is small across these tested seeds and the interval remains well above the GRPO average of $55.08$, which shows the stability of \method\ training.

\begin{table}[t]
\centering
\caption{\footnotesize Out-of-domain Avg@1 accuracy of the trained Qwen3-4B GRPO and \method\ checkpoints on two non-mathematical benchmarks, evaluated without any additional training or tuning. \#Q denotes the number of evaluated questions.}
\label{tab:ood-results}
\footnotesize
\setlength{\tabcolsep}{3.2pt}
\renewcommand{\arraystretch}{1.1}
\resizebox{\linewidth}{!}{%
\begin{tabular}{lcccc}
\toprule
\textbf{OOD Benchmark} & \textbf{\#Q} & \textbf{GRPO} & \textbf{\method} & \textbf{$\Delta$} \\
\midrule
MMLU-STEM (non-math) & $1{,}954$ & 57.06 & \textbf{60.34} & $+3.28$ \\
GPQA-Diamond & $198$ & 33.84 & \textbf{34.34} & $+0.50$ \\
\bottomrule
\end{tabular}
}
\end{table}

\noindent \textbf{Out-of-Domain Generalization.}
\method\ is an on-policy internal self-distillation method primarily developed for mathematical RLVR, and we scope our claims to this setting. To examine whether the resulting gains are confined to mathematics, we evaluate the same trained Qwen3-4B GRPO and \method\ checkpoints, without additional training or tuning, on two out-of-domain, non-mathematical benchmarks: the non-math STEM subsets of MMLU~\citep{hendrycks2020measuring} and GPQA-Diamond~\citep{rein2023gpqa}. These benchmarks assess scientific and knowledge-intensive reasoning beyond mathematics. We evaluate both benchmarks using Avg@1, with one generated response per question. Each benchmark is cast as a four-option multiple-choice task, and the model is instructed to place the final answer choice in \texttt{\textbackslash boxed\{\}}. We extract the predicted option letter and compute accuracy using exact-match comparison with the ground-truth answer.

As shown in Table~\ref{tab:ood-results}, \method\ improves over GRPO by $3.28$ points on MMLU-STEM (non-math) and by $0.50$ points on GPQA-Diamond. These results show that internal self-distillation does not reduce out-of-domain performance. The gain on MMLU-STEM also suggests that the improvement learned from mathematical RLVR can transfer to non-math STEM tasks.

\subsection{Ablation Studies}

\noindent \textbf{Effect of Logit and Attention Alignment.}
We further ablate the effects of $\mathcal{L}_{\mathrm{think}}^{(l)}$ and $\mathcal{L}_{\mathrm{attn}}^{(l)}$ by training three variants: \textsc{OISD} with logit alignment only, \textsc{OISD} with attention alignment only, and the full \textsc{OISD}. Figure~\ref{fig:component-ablation-heatmap} shows that all are effective. Logit alignment improves the macro-average from $55.08$ to $61.67$, with the largest gain on AIME24 ($32.19 \rightarrow 43.02$), while attention alignment performs slightly better overall, reaching $63.08$ on average and $46.35$ on AIME24. Combining both yields the best performance, achieving the macro-average of $64.75$. These results suggest that predictive logit alignment and evidence-routing attention alignment provide complementary supervision.

\begin{table}[t]
\centering
\caption{\footnotesize Sensitivity of \method\ to the choice of student layer on Qwen3-4B under identical training settings. AMC23 and MATH500 report Avg@16, while AIME24 and AIME25 report Avg@32. GRPO is listed as the final-layer-only reference.}
\label{tab:layer-ablation}
\footnotesize
\setlength{\tabcolsep}{3.2pt}
\renewcommand{\arraystretch}{1.1}
\resizebox{\linewidth}{!}{%
\begin{tabular}{lccccc}
\toprule
\textbf{Student layer} & \textbf{AMC23} & \textbf{MATH500} & \textbf{AIME24} & \textbf{AIME25} & \textbf{Avg.} \\
\midrule
GRPO (none) & 76.88 & 82.41 & 32.19 & 28.85 & 55.08 \\
\midrule
\rowcolor{ourscolor}
Layer $6$ (default) & 84.38 & 90.99 & \textbf{47.29} & 36.35 & 64.75 \\
Layer $26$ & 82.50 & 89.85 & 45.83 & 34.27 & 63.11 \\
Layer $35$ & \textbf{85.94} & \textbf{91.56} & 47.08 & \textbf{36.56} & \textbf{65.29} \\
Multi-layer ($6{+}26{+}35$) & 83.91 & 90.34 & 46.56 & 35.52 & 64.08 \\
\bottomrule
\end{tabular}
}
\end{table}

\noindent \textbf{Effect of the Student-Layer Selection.}
We choose the student layer using the boundary-layer criterion established in prior layer-wise analyses~\citep{tan2025bottom,yang2025unveiling,zhu2025surveylatentreasoning}, which divide Transformer layers into exploration, integration, and convergence stages and treat the exploration--integration transition as a consistent structural boundary. This criterion selects the last layer with a positive exploration signal, corresponding to layer $6$ in our setting. To assess sensitivity, we additionally evaluate layer $26$, near the end of the integration stage, layer $35$, the penultimate layer near convergence, and a multi-layer variant supervising all three. Since our focus is internal self-distillation rather than layer selection, we adopt this established criterion rather than a separate framework.

Under identical training settings on Qwen3-4B, Table~\ref{tab:layer-ablation} shows that \method\ improves over GRPO for every tested student layer, so it is not highly sensitive to the selected layer. Although layer $26$ yields the smallest gain and slightly weakens the multi-layer variant, layers $6$ and $35$ both perform strongly. Layer $35$ is easier to align because it is close to the final layer, whereas layer $6$ remains a principled choice because it enables subsequent layers to further integrate and refine the transferred knowledge. This supports the boundary-layer criterion for a relatively shallow student layer.

\noindent \textbf{Comparison between OISD and OPSD.} We also compare \method\ with the representative OPSD on-policy self-distillation framework~\cite{zhao2026self}. OPSD follows the training setting of OISD. Unlike \method, OPSD has access to privileged ground-truth solutions during training, whereas our method requires no external solution supervision. 

Table~\ref{tab:opsd-oisd-comparison} compares \method\ with OPSD on Qwen3-4B. While OPSD improves reasoning through context-based teacher--student views and already surpasses GRPO, \method\ instead performs fully internal on-policy supervision by directly distilling the model’s own final-layer logits and attention from the same rollout. Under the same Avg@$K$ evaluation protocol, OPSD achieves an average score of 56.63, whereas \method\ reaches 64.75. \method\ consistently outperforms OPSD.

\begin{table}[t]
\centering
\caption{\footnotesize \textit{Avg@K} results comparison between  GRPO, OPSD~\citep{zhao2026self}, and \method\ on Qwen3-4B. AMC23 and MATH500 report Avg@16, while AIME24 and AIME25 report Avg@32.}
\label{tab:opsd-oisd-comparison}
\footnotesize
\setlength{\tabcolsep}{3.2pt}
\renewcommand{\arraystretch}{1.1}
\resizebox{\linewidth}{!}{%
\begin{tabular}{lccccc}
\toprule
\textbf{Method} & \textbf{AMC23} & \textbf{MATH500} & \textbf{AIME24} & \textbf{AIME25} & \textbf{Avg.} \\
\midrule
GRPO & 76.88 & 82.41 & 32.19 & 28.85 & 55.08 \\
OPSD & 77.19 & 86.64 & 37.08 & 25.62 & 56.63 \\
\rowcolor{ourscolor}
\method\ (ours) & \textbf{84.38} & \textbf{90.99} & \textbf{47.29} & \textbf{36.35} & \textbf{64.75} \\
\bottomrule
\end{tabular}
}
\end{table}


\subsection{Analysis}

\begin{figure*}[t]
\centering
\includegraphics[width=\textwidth]{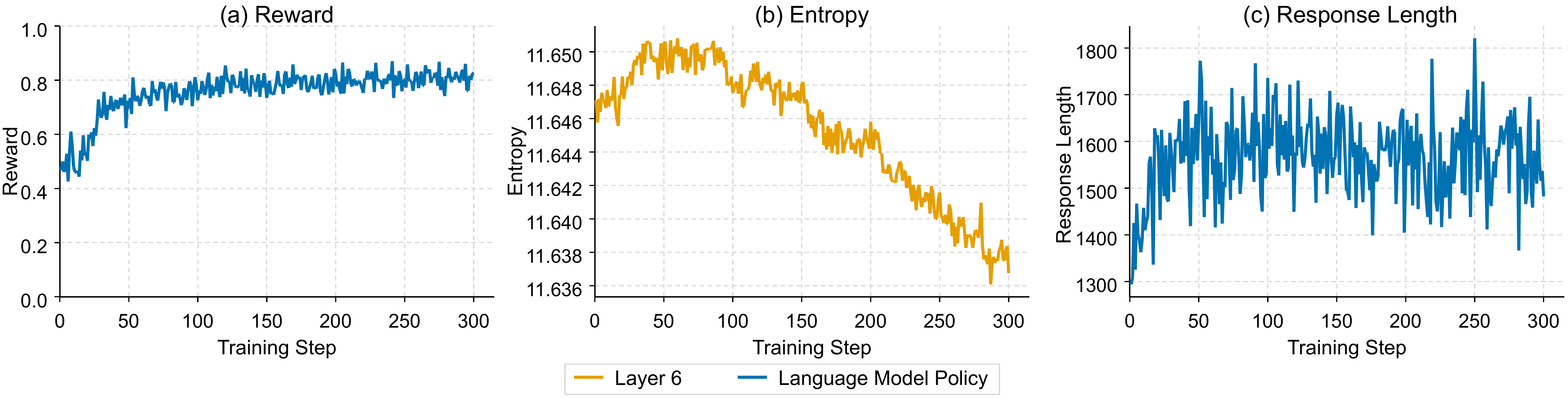}
\caption{Training dynamics of \textsc{OISD} over the full RL run on Qwen3-4B. (a) Average rollout reward of the final-layer acting policy increases steadily across training. (b) Response-token entropy of the middle-layer readout (layer 6). (c) The response length of generated trajectories grows as the policy develops longer reasoning chains.}
\label{fig:training-dynamics}
\end{figure*}

\noindent \textbf{Training Dynamics.} Figure~\ref{fig:training-dynamics} summarizes the training dynamics of \textsc{OISD} with the Qwen3-4B backbone. We track reward, layer-6 entropy, and response length to analyze the effect of OISD. The reward steadily improves during training, while the entropy increases during the early stage of training, suggesting broader policy exploration and a larger reasoning search space induced by OISD. The response length grows from roughly 1300 to 1600 tokens before stabilizing with fluctuations.

\begin{figure}[t]
\centering
\includegraphics[width=\linewidth]{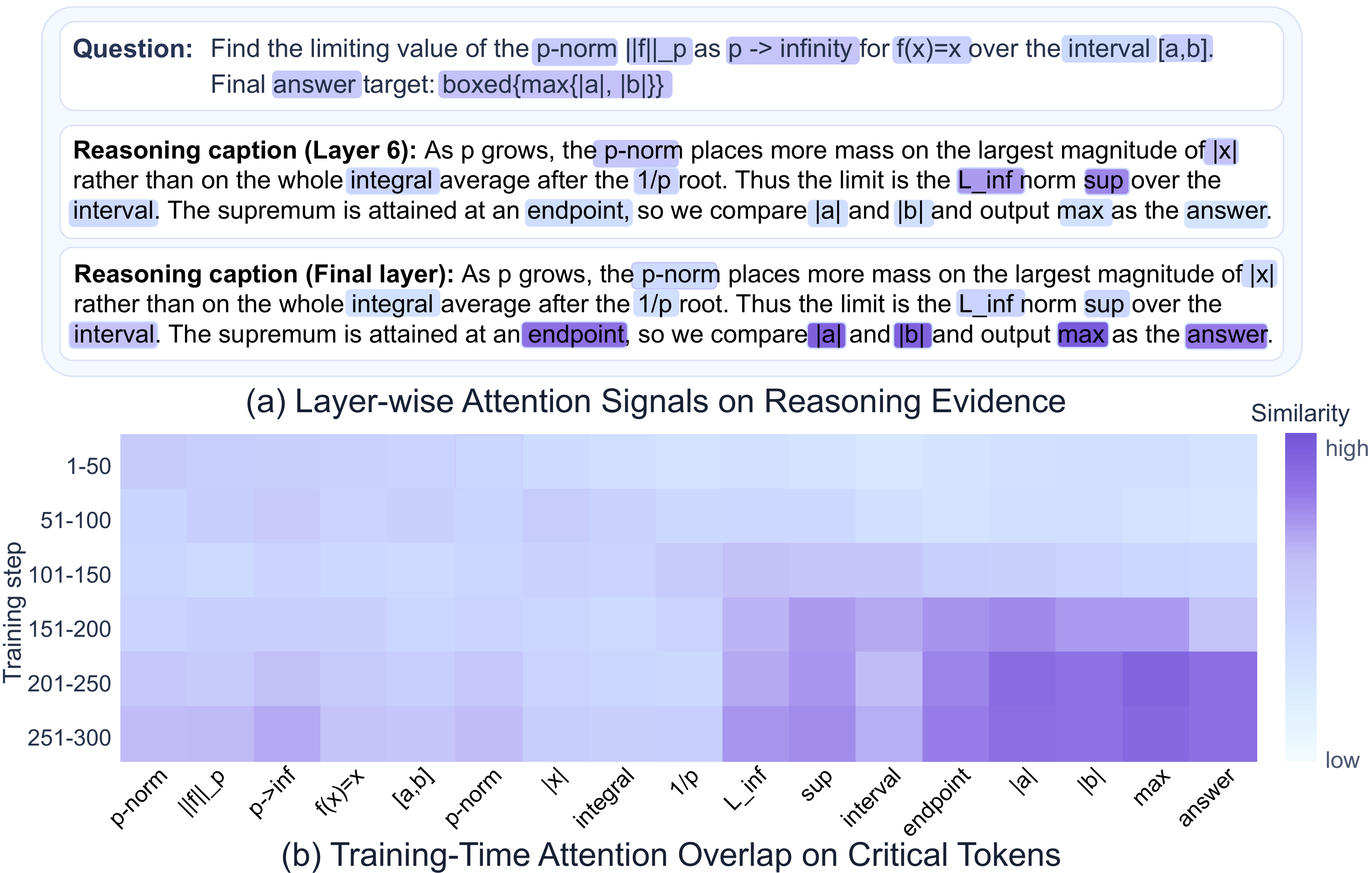}
\caption{
Attention visualization during \textsc{OISD} training. (a) Attention highlights from the intermediate and final layers on the same reasoning trajectory, where highlighted tokens support the final answer. (b) Similarity between intermediate- and final-layer attention on critical tokens across training blocks, with darker colors indicating stronger alignment (higher similarity). 
}
\label{fig:attention-visualization}
\end{figure}

\begin{figure}[t]
\centering
\maybeincludegraphics[width=\linewidth]{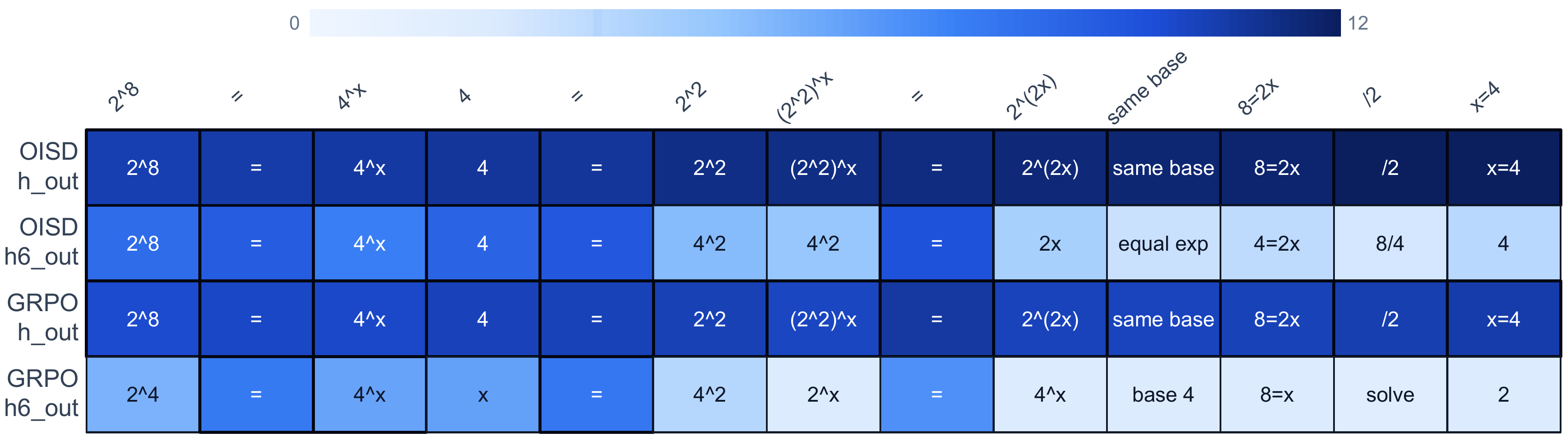}
\caption{Logit-lens comparison between \method\ and GRPO on a reasoning trace. Each row shows the top token decoded from the final or layer-$6$ hidden state. Bold outlines indicate agreement between layer-$6$ and final-layer predictions. \method\ exhibits earlier alignment on reasoning-critical tokens, whereas GRPO   produces more local or incomplete intermediate predictions.}
\label{fig:logit-lens-layer-compare}
\end{figure}

\noindent \textbf{Qualitative Attention Visualization.} We explore how attention alignment shapes intermediate-layer reasoning in Figure~\ref{fig:attention-visualization}. In Figure~\ref{fig:attention-visualization}(a), we use a fixed reasoning example and mark the words that provide critical evidence for the final answer. For layer 6 and the final layer, we average causal attention from
  selected late-reasoning query positions over heads, merge subwords into word-level scores, and normalize within each layer. Darker color highlights indicate stronger attention
  to the marked evidence words. The final layer places stronger attention on answer-supporting evidence, including the limiting norm, supremum, endpoint comparison, and final maximum operation, while
  layer 6 shows a broader but partially overlapping pattern. 
  
  Figure~\ref{fig:attention-visualization}(b) tracks this alignment during training. We compute attention
  Jensen--Shannon divergence between layer 6 and the final layer and convert it into a normalized agreement score, where higher values indicate stronger overlap. Columns
  correspond to the critical-token groups in Figure~\ref{fig:attention-visualization}(a). Darker colors in later stages indicate that attention alignment gradually moves the
  intermediate layer toward the final layer's evidence-routing pattern.

\noindent \textbf{Qualitative Logit-Lens Visualization.} Figure~\ref{fig:logit-lens-layer-compare} presents a logit-lens comparison between GRPO and \textsc{OISD} on a representative reasoning trace using the Qwen3-4B backbone. Each row shows the top decoded token from either the final layer or the intermediate layer-$6$ hidden state, while bold outlines denote agreement between layer-$6$ and final-layer predictions. \textsc{OISD} exhibits substantially earlier alignment between intermediate and final predictions on reasoning-critical tokens, indicating that the intermediate layer develops more globally consistent reasoning behaviors during generation. In contrast, GRPO often produces more local, fragmented, or incomplete intermediate predictions.

\section{Conclusion}

We introduce \textit{on-policy internal self-distillation}, a new paradigm for reasoning RL that leverages the model’s own internal computation as supervision while preserving a single acting policy for rollout and optimization. Building on this idea, we propose \textsc{OISD}, which distills the model’s final-layer computation into an intermediate layer through two complementary signals: logit alignment for \emph{how to think} and attention alignment for \emph{where to look}.  \textsc{OISD} preserves policy consistency and provides persistent internal supervision for learning improved intermediate representations. Experiments on Qwen3-4B demonstrate substantial gains over strong reasoning RL baselines, highlighting on-policy internal self-distillation as a promising direction for reasoning-oriented LLM post-training.

\clearpage
\newpage

\section*{Limitations}

Limited computational resources constrained our exploration of broader model scales and architectures. Nevertheless, this work provides a promising perspective on on-policy internal self-distillation for LLM reasoning. Our experiments mainly focus on mathematical reasoning with verifiable answers, and future work should investigate whether the proposed framework generalizes to more complex reasoning tasks requiring longer reasoning chains or different forms of supervision. In addition, we do not conduct an in-depth study of layer selection. Extending internal supervision to multiple layers could be an interesting direction for future research.


\bibliography{custom}


\appendix
\section{Appendix}

\subsection{Training Dynamics}

\begin{figure}[htbp]
\centering
\maybeincludegraphics[width=\linewidth]{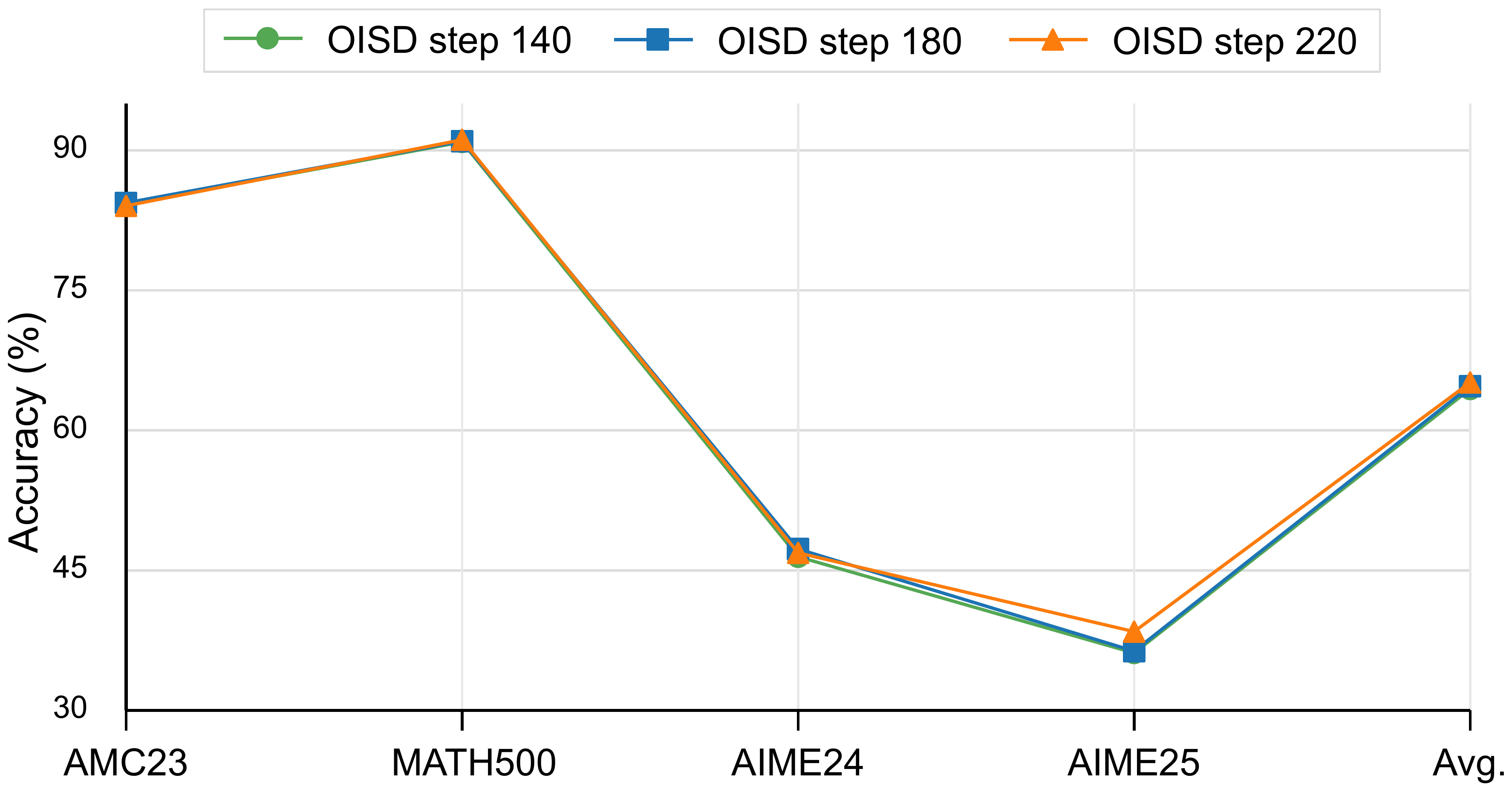}
\caption{Checkpoint progression for Qwen3-4B \method\ with logit and attention alignment. Results are reported with Avg@16 on AMC23 and MATH500, Avg@32 on AIME24 and AIME25, and the four-benchmark average.}
\label{fig:checkpoint-progression}
\end{figure}

As shown in Figure~\ref{fig:checkpoint-progression}, Qwen3-4B \method\ steadily improves as training progresses. Performance increases across most benchmarks from early to later checkpoints, leading to a higher overall average. In the late stage, the results stabilize within a narrow range.

\subsection{Training Details}
\label{sec:appendix-training-details}

All experiments are conducted on four NVIDIA A100 or H200 GPUs.
Since reinforcement fine-tuning for long-form reasoning is computationally intensive, we utilize FlashAttention, FSDP-2 sharding, gradient checkpointing, dynamic batching by token length, and vLLM rollouts with chunked prefill to improve training throughput and memory efficiency.
The implementation is built with PyTorch 2.6.0+cu124, Transformers 4.56.0, vLLM 0.8.5, Ray 2.47.1, and PEFT 0.18.1.

To ensure consistency and comparability across experiments, we adopt the same on-policy training pipeline for GRPO, OPSD and \method. 
We follow the ~\citep{tan2025bottom} for all shared settings, including the \texttt{deepmath-5k} training subset, a prompt batch size of $128$, and $8$ rollouts per prompt.
We use AdamW with a learning rate of $1\!\times\!10^{-6}$, the GRPO advantage estimator, token-mean loss aggregation, and a symmetric policy clipping window of $\epsilon=0.2$.

For \method, the rollout policy is always read from the final Transformer layer, while the shadow student is fixed at layer $l=6$.
We set the two internal loss weights to $\lambda_{\mathrm{think}}{=}1.0$ and $\lambda_{\mathrm{attn}}{=}0.1$. 

\begin{table}[htbp]
\centering
\caption{\footnotesize Training cost of GRPO and \method\ on Qwen3-4B under identical data, rollout configurations, hardware, and numbers of training steps. Throughput is reported in tokens/s/GPU.}
\label{tab:efficiency}
\footnotesize
\setlength{\tabcolsep}{3.2pt}
\renewcommand{\arraystretch}{1.1}
\resizebox{\linewidth}{!}{%
\begin{tabular}{lccc}
\toprule
\textbf{Qwen3-4B ($4\times$A100-80GB)} & \textbf{GRPO} & \textbf{\method} & \textbf{Diff.} \\
\midrule
Time per step & 273.0 s & 310.4 s & $+13.7\%$ \\
\quad Generation / rollout & 126.0 s & 124.9 s & $\approx 0\%$ \\
\quad Actor update & 116.3 s & 153.0 s & $+31.6\%$ \\
Throughput & $1{,}574$ & $1{,}379$ & $-12.4\%$ \\
Peak memory per GPU & 28.9 GB & 31.9 GB & $+10.7\%$ \\
Total training time (300 steps) & 22.8 h & 25.9 h & $+3.1$ h \\
\bottomrule
\end{tabular}
}
\end{table}

\noindent \textbf{Computational Cost.}
We report the wall-clock training time, throughput, and peak GPU memory of \method\ and GRPO under identical data, rollout configurations, hardware, and numbers of training steps. As shown in Table~\ref{tab:efficiency}, for Qwen3-4B \method\ incurs a $13.7\%$ increase in wall-clock time and a $10.7\%$ increase in peak GPU memory relative to GRPO, accompanied by a $12.4\%$ reduction in throughput. The overhead is concentrated in the actor update ($+31.6\%$), while generation is essentially unchanged, since the internal alignment losses are computed only during the policy update and the rollout is still produced by the final layer alone. For reference, BuPO~\citep{tan2025bottom} reports 13.22 hours for a complete Qwen3-4B reinforcement-learning run and an additional 209.25 seconds for internal-policy analysis. For Qwen3-8B, \method\ takes 434.5 seconds per step, reaches 1,474 tokens/s/GPU, and uses 51.3 GB peak memory per GPU on $4\times$H200 GPUs, giving a total training time of about 36.2 hours for 300 steps. We do not have a GRPO run under the same hardware and configuration at this scale under our current computational resource constraints.

\subsection{Gradients of the Think and Attention Losses}
\label{sec:appendix-gradients}

\subsubsection{Gradients of the Think loss} 

Recall that, with temperature $\tau$, the layer-$l$ shadow student and the final-layer teacher distributions are
\[
\begin{aligned}
\mathbf{z}^{l}_{t}
&=
\mathrm{LN}(\mathbf{h}^{l}_{t})\mathbf{E}_{\mathrm{u}}^\top, \\
\mathbf{z}^{L}_{t}
&=
\mathrm{LN}(\mathbf{h}^{L}_{t})\mathbf{E}_{\mathrm{u}}^\top, \\
\mathbf{p}^{l,\tau}_{t}
&=
\mathrm{Softmax}(\mathbf{z}^{l}_{t}/\tau), \\
\mathbf{q}^{L,\tau}_{t}
&=
\operatorname{sg}
\!\left[
\mathrm{Softmax}(\mathbf{z}^{L}_{t}/\tau)
\right],
\end{aligned}
\]
where $\operatorname{sg}(\cdot)$ denotes stop-gradient on the detached final-layer teacher. Define the mixture distribution
\[
\mathbf{m}^{l,L,\tau}_{t}
=
\frac{1}{2}
\left(
\mathbf{p}^{l,\tau}_{t}
+
\mathbf{q}^{L,\tau}_{t}
\right).
\]

The token-level Jensen--Shannon divergence is
\[
\begin{aligned}
\operatorname{JS}
\left(
\mathbf{p}^{l,\tau}_{t},
\mathbf{q}^{L,\tau}_{t}
\right)
& =
\frac{1}{2}
\operatorname{KL}
\left(
\mathbf{p}^{l,\tau}_{t}
\middle\|
\mathbf{m}^{l,L,\tau}_{t}
\right) \\
& +
\frac{1}{2}
\operatorname{KL}
\left(
\mathbf{q}^{L,\tau}_{t}
\middle\|
\mathbf{m}^{l,L,\tau}_{t}
\right).
\end{aligned}
\]
Expanding over vocabulary index $i$ gives
\[
\operatorname{JS}
=
\frac{1}{2}\sum_i
p_i
\log
\frac{p_i}{m_i}
+
\frac{1}{2}\sum_i
q_i
\log
\frac{q_i}{m_i},
\]
where $p_i=\mathbf{p}^{l,\tau}_{t,i}$, $q_i=\mathbf{q}^{L,\tau}_{t,i}$, and $m_i=\mathbf{m}^{l,L,\tau}_{t,i}$.

Since the teacher distribution is detached via the stop-gradient and 
\[
m_i=\frac{1}{2}(p_i+q_i),
\qquad
\frac{\partial m_i}{\partial p_i}=\frac{1}{2},
\]
the derivative with respect to $p_i$ becomes
\[
\begin{aligned}
\frac{\partial \operatorname{JS}}{\partial p_i}
&=
\frac{1}{2}
\left(
\log\frac{p_i}{m_i}
+
1
-
\frac{p_i}{2m_i}
\right)
-
\frac{q_i}{4m_i} \\
&=
\frac{1}{2}\log\frac{p_i}{m_i}
+
\frac{1}{2}
-
\frac{p_i+q_i}{4m_i}.
\end{aligned}
\]
Using $m_i=\frac{1}{2}(p_i+q_i)$, the last two terms cancel, yielding
\[
\boxed{
\frac{\partial \operatorname{JS}}{\partial p_i}
=
\frac{1}{2}\log\frac{p_i}{m_i}
}.
\]

Define
\[
\mathbf{g}_{t}
=
\frac{1}{2}
\log
\frac{
\mathbf{p}^{l,\tau}_{t}
}{
\mathbf{m}^{l,L,\tau}_{t}
}.
\]

Using the chain rule, we have
\[
\frac{\partial \mathcal{L}_{\mathrm{think}}^{(l)}}
{\partial \mathbf{z}^{l}_{t}}
=
A_t
\left(
\frac{\partial \mathbf{p}^{l,\tau}_{t}}
{\partial \mathbf{z}^{l}_{t}}
\right)^\top
\frac{\partial \operatorname{JS}}
{\partial \mathbf{p}^{l,\tau}_{t}} .
\]
From the previous derivation,
\[
\frac{\partial \operatorname{JS}}
{\partial \mathbf{p}^{l,\tau}_{t}}
=
\mathbf{g}_{t},
\qquad
\mathbf{g}_{t}
=
\frac{1}{2}
\log
\frac{\mathbf{p}^{l,\tau}_{t}}
{\mathbf{m}^{l,L,\tau}_{t}} .
\]
The softmax Jacobian is
\[
\frac{\partial \mathbf{p}^{l,\tau}_{t}}
{\partial \mathbf{z}^{l}_{t}}
=
\frac{1}{\tau}
\left[
\operatorname{diag}(\mathbf{p}^{l,\tau}_{t})
-
\mathbf{p}^{l,\tau}_{t}
(\mathbf{p}^{l,\tau}_{t})^\top
\right].
\]
Applying the chain rule together with the softmax Jacobian gives
\[
\begin{aligned}
\frac{\partial \mathcal{L}_{\mathrm{think}}^{(l)}}
{\partial \mathbf{z}^{l}_{t}}
&=
A_t
\left(
\frac{\partial \mathbf{p}^{l,\tau}_{t}}
{\partial \mathbf{z}^{l}_{t}}
\right)^\top
\frac{\partial \operatorname{JS}}
{\partial \mathbf{p}^{l,\tau}_{t}} \\
&=
\frac{A_t}{\tau}
\left[
\operatorname{diag}(\mathbf{p}^{l,\tau}_{t})
-
\mathbf{p}^{l,\tau}_{t}
(\mathbf{p}^{l,\tau}_{t})^\top
\right]
\mathbf{g}_{t},
\end{aligned}
\]
where
\[
\mathbf{g}_{t}
=
\frac{1}{2}
\log
\frac{
\mathbf{p}^{l,\tau}_{t}
}{
\mathbf{m}^{l,L,\tau}_{t}
}.
\]

Expanding the matrix-vector product yields
\[
\begin{aligned}
\frac{\partial \mathcal{L}_{\mathrm{think}}^{(l)}}
{\partial \mathbf{z}^{l}_{t}}
&=
\frac{A_t}{\tau}
\left[
\operatorname{diag}(\mathbf{p}^{l,\tau}_{t})\mathbf{g}_{t}
-
\mathbf{p}^{l,\tau}_{t}
(\mathbf{p}^{l,\tau}_{t})^\top
\mathbf{g}_{t}
\right].
\end{aligned}
\]

For the first term,
\[
\operatorname{diag}(\mathbf{p}^{l,\tau}_{t})\mathbf{g}_{t}
=
\mathbf{p}^{l,\tau}_{t}\odot \mathbf{g}_{t},
\]
while for the second term,
\[
(\mathbf{p}^{l,\tau}_{t})^\top \mathbf{g}_{t}
=
\left\langle
\mathbf{p}^{l,\tau}_{t},
\mathbf{g}_{t}
\right\rangle
\]
is a scalar, giving
\[
\mathbf{p}^{l,\tau}_{t}
(\mathbf{p}^{l,\tau}_{t})^\top
\mathbf{g}_{t}
=
\mathbf{p}^{l,\tau}_{t}
\left\langle
\mathbf{p}^{l,\tau}_{t},
\mathbf{g}_{t}
\right\rangle .
\]

Therefore,
\[
\begin{aligned}
\frac{\partial \mathcal{L}_{\mathrm{think}}^{(l)}}
{\partial \mathbf{z}^{l}_{t}}
&=
\frac{A_t}{\tau}
\left[
\mathbf{p}^{l,\tau}_{t}\odot \mathbf{g}_{t}
-
\mathbf{p}^{l,\tau}_{t}
\left\langle
\mathbf{p}^{l,\tau}_{t},
\mathbf{g}_{t}
\right\rangle
\right].
\end{aligned}
\]

Since the inner product term is scalar-valued,
\[
\mathbf{p}^{l,\tau}_{t}
\left\langle
\mathbf{p}^{l,\tau}_{t},
\mathbf{g}_{t}
\right\rangle
=
\mathbf{p}^{l,\tau}_{t}
\odot
\left(
\left\langle
\mathbf{p}^{l,\tau}_{t},
\mathbf{g}_{t}
\right\rangle
\mathbf{1}
\right),
\]
Thus,
\[
\boxed{
\frac{\partial \mathcal{L}_{\mathrm{think}}^{(l)}}
{\partial \mathbf{z}^{l}_{t}}
=
\frac{A_t}{\tau}
\left[
\mathbf{p}^{l,\tau}_{t}
\odot
\left(
\mathbf{g}_{t}
-
\left\langle
\mathbf{p}^{l,\tau}_{t},
\mathbf{g}_{t}
\right\rangle
\mathbf{1}
\right)
\right]
}.
\]

Since
\[
\mathbf{z}^{l}_{t}
=
\mathrm{LN}(\mathbf{h}^{l}_{t})
\mathbf{E}_{\mathrm{u}}^\top,
\]
the chain rule gives
\[
\boxed{
\frac{\partial \mathcal{L}_{\mathrm{think}}^{(l)}}
{\partial \mathbf{h}^{l}_{t}}
=
J_{\mathrm{LN}}(\mathbf{h}^{l}_{t})^\top
\mathbf{E}_{\mathrm{u}}
\frac{\partial \mathcal{L}_{\mathrm{think}}^{(l)}}
{\partial \mathbf{z}^{l}_{t}}
}
\]
where $J_{\mathrm{LN}}(\mathbf{h}^{l}_{t})$ denotes the Jacobian of layer normalization. The clipped advantage
\[
A_t=\operatorname{clip}(A_t,-c,c)
\]
acts as a signed scaling factor: positive-advantage tokens pull the shadow student toward the detached final-layer teacher, while negative-advantage tokens reverse the gradient direction to suppress poor reasoning trajectories.

\subsubsection{Gradients of  Attention Loss}

Let
\[
\mathbf p_{t,h}^{l}
=
\tilde a_{t,h}^{l},
\qquad
\mathbf p_{t,h}^{L}
=
\sg(\tilde a_{t,h}^{L}),
\]
where $\mathbf p_{t,h}^{l}$ denotes the sampled attention distribution of the shadow layer and $\mathbf p_{t,h}^{L}$ denotes the detached final-layer teacher distribution over the sampled causal set $\tilde{\mathcal C}_t$.

For decoding step $t$ and attention head $h$, the attention-alignment loss is
\[
\mathcal L_{\mathrm{attn}}^{(l)}
=
\frac{A_t}{H}
\operatorname{JS}
\left(
\mathbf p_{t,h}^{l},
\mathbf p_{t,h}^{L}
\right).
\]

Define the mixture distribution
\[
\mathbf m_{t,h}
=
\frac{1}{2}
\left(
\mathbf p_{t,h}^{l}
+
\mathbf p_{t,h}^{L}
\right).
\]

Using the same JS-divergence derivation as before,
\[
\frac{
\partial
\operatorname{JS}
(
\mathbf p_{t,h}^{l},
\mathbf p_{t,h}^{L}
)
}{
\partial \mathbf{p}_{t,h}^{l}(j)
}
=
\frac{1}{2}
\log
\frac{
\mathbf{p}_{t,h}^{l}(j)
}{
\mathbf{m}_{t,h}(j)
}.
\]

Therefore,
\[
\frac{
\partial
\mathcal L_{\mathrm{attn}}^{(l)}
}{
\partial \mathbf{p}_{t,h}^{l}(j)
}
=
\frac{A_t}{2H}
\log
\frac{
\mathbf{p}_{t,h}^{l}(j)
}{
\mathbf{m}_{t,h}(j)
}.
\]

Let the sampled attention logits at layer $l$ be
\[
s_{t,h}^{l}(j)
=
\frac{
(\mathbf q_{t,h}^{l})^\top
\mathbf k_{j,h}^{l}
}{
\sqrt{d_h}
},
\qquad
j\in\tilde{\mathcal C}_t,
\]
with
\[
p_{t,h}^{l}(j)
=
\mathrm{Softmax}
(
\mathbf s_{t,h}^{l}
)_j.
\]

Using the softmax Jacobian,
\[
\frac{
\partial p_{t,h}^{l}(j)
}{
\partial s_{t,h}^{l}(i)
}
=
p_{t,h}^{l}(j)
\left(
\mathbf 1_{i=j}
-
p_{t,h}^{l}(i)
\right),
\]
the gradient with respect to the sampled attention logits becomes
\[
\frac{
\partial
\mathcal L_{\mathrm{attn}}^{(l)}
}{
\partial s_{t,h}^{l}(i)
}
=
A_t\,
p_{t,h}^{l}(i)
\left(
g_{t,h}(i)
-
\sum_{j\in\tilde{\mathcal C}_t}
p_{t,h}^{l}(j)
g_{t,h}(j)
\right),
\]
where
\[
g_{t,h}(i)
=
\frac{1}{2H}
\log
\frac{
p_{t,h}^{l}(i)
}{
m_{t,h}(i)
}.
\]

Equivalently, in vector form,
\[
\boxed{
\frac{
\partial
\mathcal L_{\mathrm{attn}}^{(l)}
}{
\partial \mathbf s_{t,h}^{l}
}
=
A_t\,
\mathbf p_{t,h}^{l}
\odot
\left(
\mathbf g_{t,h}
-
\langle
\mathbf p_{t,h}^{l},
\mathbf g_{t,h}
\rangle
\mathbf 1
\right)
}
\]
with
\[
\mathbf g_{t,h}
=
\frac{1}{2H}
\log
\frac{
\mathbf p_{t,h}^{l}
}{
\mathbf m_{t,h}
}.
\]

Since the sampled attention logits are defined as
\[
s_{t,h}^{l}(j)
=
\frac{
(\mathbf q_{t,h}^{l})^\top
\mathbf k_{j,h}^{l}
}{
\sqrt{d_h}
},
\qquad
j\in\tilde{\mathcal C}_t,
\]
their partial derivatives are
\[
\frac{
\partial s_{t,h}^{l}(j)
}{
\partial \mathbf q_{t,h}^{l}
}
=
\frac{
\mathbf k_{j,h}^{l}
}{
\sqrt{d_h}
},
\qquad
\frac{
\partial s_{t,h}^{l}(j)
}{
\partial \mathbf k_{j,h}^{l}
}
=
\frac{
\mathbf q_{t,h}^{l}
}{
\sqrt{d_h}
}.
\]

Applying the chain rule, the gradient with respect to the query vector becomes
\[
\boxed{
\frac{
\partial
\mathcal L_{\mathrm{attn}}^{(l)}
}{
\partial \mathbf q_{t,h}^{l}
}
=
\frac{1}{\sqrt{d_h}}
\sum_{j\in\tilde{\mathcal C}_t}
\frac{
\partial
\mathcal L_{\mathrm{attn}}^{(l)}
}{
\partial s_{t,h}^{l}(j)
}
\mathbf k_{j,h}^{l}
}
\]

Similarly, for each sampled key position $j$,
\[
\boxed{
\frac{
\partial
\mathcal L_{\mathrm{attn}}^{(l)}
}{
\partial \mathbf k_{j,h}^{l}
}
=
\frac{1}{\sqrt{d_h}}
\frac{
\partial
\mathcal L_{\mathrm{attn}}^{(l)}
}{
\partial s_{t,h}^{l}(j)
}
\mathbf q_{t,h}^{l}
}
\]

The corresponding compact vector forms are
\[
\boxed{
\frac{
\partial
\mathcal L_{\mathrm{attn}}^{(l)}
}{
\partial \mathbf q_{t,h}^{l}
}
=
\frac{1}{\sqrt{d_h}}
(\mathbf K_{\tilde{\mathcal C}_t,h}^{l})^\top
\frac{
\partial
\mathcal L_{\mathrm{attn}}^{(l)}
}{
\partial \mathbf s_{t,h}^{l}
}
}
\]

and
\[
\boxed{
\frac{
\partial
\mathcal L_{\mathrm{attn}}^{(l)}
}{
\partial
\mathbf K_{\tilde{\mathcal C}_t,h}^{l}
}
=
\frac{1}{\sqrt{d_h}}
\frac{
\partial
\mathcal L_{\mathrm{attn}}^{(l)}
}{
\partial \mathbf s_{t,h}^{l}
}
(\mathbf q_{t,h}^{l})^\top
}
\]

\subsubsection{Gradient Norm Visualization}

\begin{figure}[htbp]
\centering
\maybeincludegraphics[width=\linewidth]{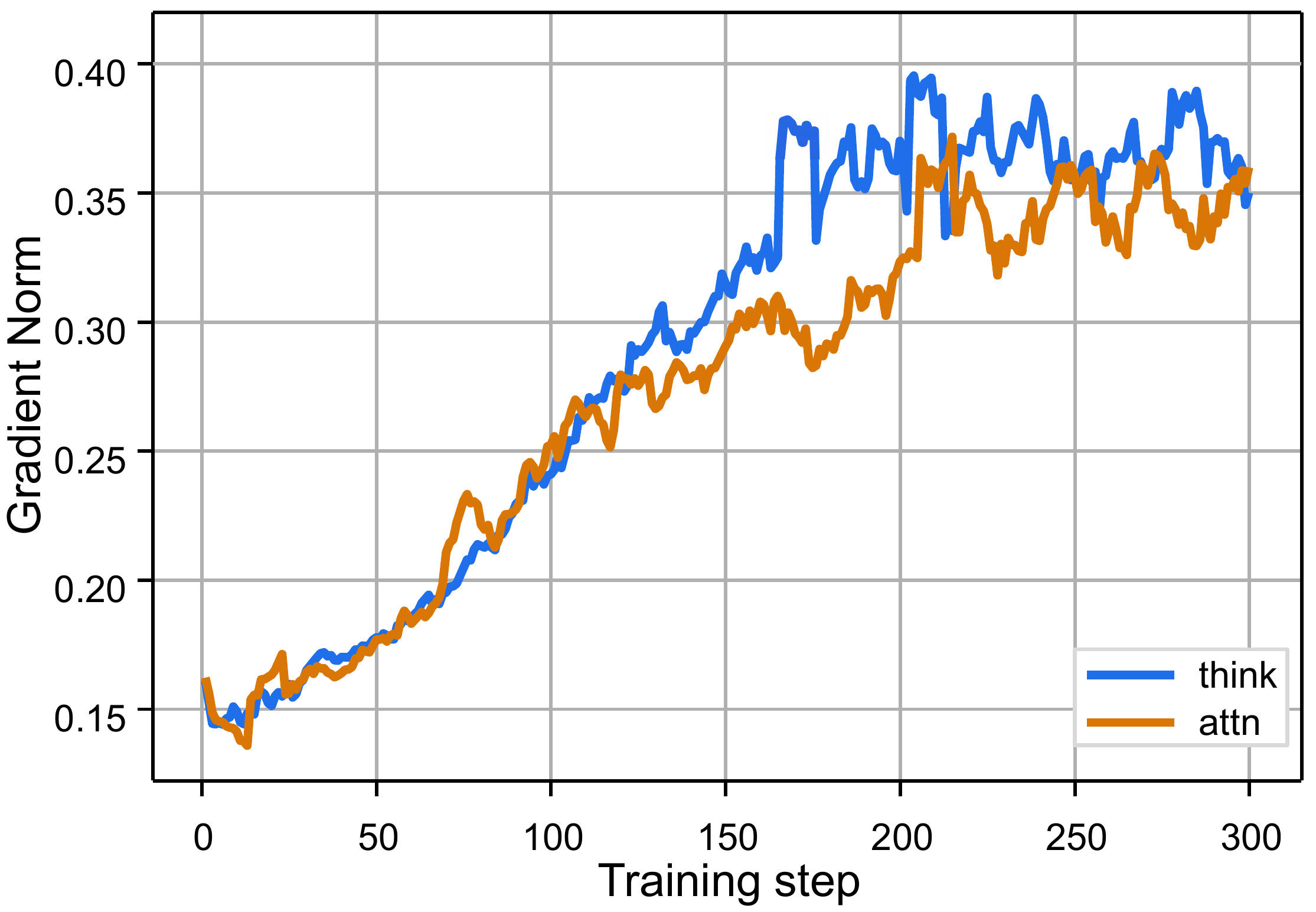}
\caption{Gradient norm for the think and attention alignment losses.}
\label{fig:loss-gradient-comparison}
\end{figure}

Figure~\ref{fig:loss-gradient-comparison} shows the actor gradient norms induced by the two internal alignment losses. The think loss $\mathcal{L}_{\mathrm{think}}$ transfers token-level beliefs from the final-layer teacher to the intermediate-layer student, while the attention loss $\mathcal{L}_{\mathrm{attn}}$ transfers evidence-routing patterns over the sampled causal context.

As shown in Figure~\ref{fig:loss-gradient-comparison}, both losses produce stable and progressively stronger gradient signals throughout training. The increasing gradient norm suggests that \textsc{OISD} gradually strengthens the interaction between intermediate and final-layer reasoning states. As RL training improves the final-layer policy, the resulting teacher--student mismatch provides increasingly informative internal supervision, leading to stronger latent-space supervision on intermediate representations.

\subsection{Sample Prediction Comparison}
\label{sec:appendix-sample-prediction}
For qualitative comparison, we inspect the reasoning traces of GRPO, OPSD, and \method\ on the same AIME~2024 tetrahedron problem. The traces and per-problem statistics are taken from the corresponding AIME~2024 Avg@$32$ evaluation dump used in the main comparison.

\begin{figure*}[tp]
\centering
\maybeincludegraphics[width=\textwidth]{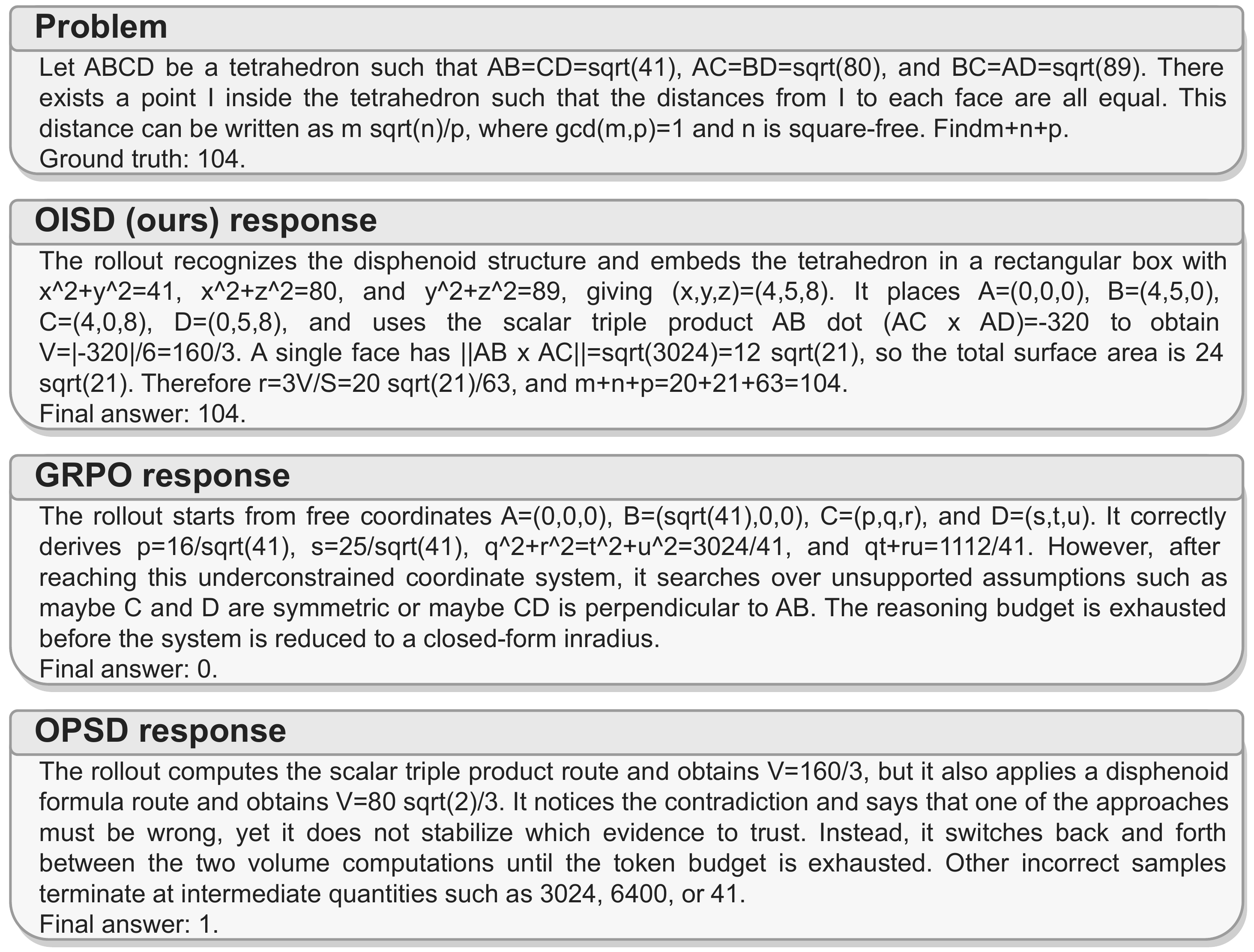}
\caption{Reasoning comparison on the same AIME~2024 tetrahedron problem. The top box states the problem and ground truth; each column shows the first sampled rollout for one method together with its Avg@$32$ pass rate on this problem.}
\label{fig:sample-reasoning-comparison}
\end{figure*}
\paragraph{Problem.} \emph{Let $ABCD$ be a tetrahedron such that $AB{=}CD{=}\sqrt{41}$, $AC{=}BD{=}\sqrt{80}$, and $BC{=}AD{=}\sqrt{89}$. There exists a point $I$ inside the tetrahedron such that the distances from $I$ to each of the faces of the tetrahedron are all equal. This distance can be written in the form $\tfrac{m\sqrt{n}}{p}$, where $m,n,p$ are positive integers, $m$ and $p$ are relatively prime, and $n$ is not divisible by the square of any prime. Find $m+n+p$.} \emph{Ground truth:} 104.

As shown in Figure~\ref{fig:sample-reasoning-comparison}, \method's first rollout closes the reasoning chain by recognizing the disphenoid structure, computing $V{=}\tfrac{160}{3}$ and $r{=}\tfrac{20\sqrt{21}}{63}$, and returning 104.
By contrast, the GRPO rollout derives several correct coordinate constraints but then searches over unsupported symmetry assumptions and terminates with 0.
OPSD obtains the useful scalar-triple-product volume but also produces a conflicting formula-based volume, repeatedly revisits the disagreement, and finally emits the answer 1.

\subsection{Additional Ablations}
\label{sec:appendix-ablations}

\paragraph{Loss weight $\lambda_{\mathrm{think}}$ in the think-only setting.}
The logit-alignment weight $\lambda_{\mathrm{think}}$ controls the strength of the internal teacher signal in the think-only setting.
A smaller value weakens the pressure from the final-layer distribution, while a larger value makes the shadow layer follow the teacher more aggressively.
We set $\lambda_{\mathrm{think}}\in\{0.5,1.0,2.0\}$ with the attention branch disabled and report the comparison in Table~\ref{tab:lambda-ablation}. 
\begin{table}[htbp]
\centering
\caption{Sensitivity of the think-only (logit-only) \method\ variant to $\lambda_{\mathrm{think}}$ on Qwen3-4B. AMC23 and MATH500 report Avg@16, AIME24 and AIME25 report Avg@32.}
\label{tab:lambda-ablation}
\footnotesize
\setlength{\tabcolsep}{3.5pt}
\renewcommand{\arraystretch}{1.05}
\begin{tabular}{lccccc}
\toprule
$\lambda_{\mathrm{think}}$ & \textbf{AMC23} & \textbf{MATH500} & \textbf{AIME24} & \textbf{AIME25} & \textbf{Avg.} \\
\midrule
$0.5$  & 82.50 & 87.39 & 41.25 & 30.52 & 60.41 \\
\rowcolor{ourscolor}
$1.0$  & 81.25 & 89.80 & 43.02 & 32.60 & 61.67 \\
$2.0$  & 81.09 & 86.36 & 40.83 & 31.56 & 59.96 \\
\bottomrule
\end{tabular}
\end{table}

With $\lambda_{\mathrm{think}}{=}0.5$, the auxiliary logit alignment already improves the internal-student variant, but the signal appears insufficient to fully align the intermediate layer with the final-layer teacher.
Increasing the weight to $1.0$ gives the best average performance, making the think loss provides useful internal supervision while remaining compatible with the on-policy GRPO update.

When the weight is increased to $2.0$, the auxiliary objective becomes too strong and performance drops. This indicates that logit alignment should guide the intermediate layer without competing with the on-policy update. We therefore use $\lambda_{\mathrm{think}}{=}1.0$ as the default think-loss weight in the main experiments. We choose the attention-loss weight by the same criterion and set $\lambda_{\mathrm{attn}}{=}0.1$.


\subsection{Additional Results}
\label{sec:appendix-additional-results}

\begin{table}[htbp]
\centering
\caption{\footnotesize Correct-only equation-level diversity (Div-Equ, higher is better) on Qwen3-4B, following~\citet{lochab2026uniform}. For each shared valid problem we sample $m=16$ correct responses per method.}
\label{tab:diversity}
\footnotesize
\setlength{\tabcolsep}{4pt}
\renewcommand{\arraystretch}{1.1}
\begin{tabular}{lccc}
\toprule
\textbf{Benchmark} & \textbf{GRPO} & \textbf{\method} & \textbf{$\Delta$} \\
\midrule
AMC23   & 0.184 & \textbf{0.186} & $+0.002$ \\
MATH500 & 0.128 & \textbf{0.140} & $+0.012$ \\
AIME24  & 0.220 & \textbf{0.232} & $+0.012$ \\
AIME25  & 0.166 & \textbf{0.174} & $+0.008$ \\
\midrule
\rowcolor{ourscolor}
Avg.    & 0.175 & \textbf{0.183} & $+0.009$ \\
\bottomrule
\end{tabular}
\end{table}

\noindent \textbf{Solution Diversity.}
Higher Pass@$K$ could in principle be obtained from a narrower set of correct solutions, so we also examine whether \method\ preserves solution diversity. Following UCPO~\citep{lochab2026uniform}, we measure correct-only equation-level diversity using Div-Equ, which computes the fraction of mathematical expressions in a correct rollout that are unique relative to the other correct rollouts for the same problem. To make a fair comparison, we sample the same number of correct responses for both methods, using $m=16$ correct responses per method for each shared valid problem. As reported in Table~\ref{tab:diversity}, \method\ achieves higher Div-Equ on all four benchmarks, with an average gain of $0.009$. These results show that \method\ improves accuracy without reducing the equation-level diversity of correct solutions.

\subsection{Information About Use of AI Assistants}

ChatGPT was used for language editing, manuscript polishing, and assistance with part of the equation derivations of JS. All outputs were reviewed and verified by the authors.

\end{document}